\documentclass[conference, 9pt]{IEEEtran}
\usepackage{geometry}
\usepackage{acronym}
\usepackage{algorithm}
\usepackage{algorithmic}
\usepackage{amsmath,amssymb,amsfonts}
\usepackage{booktabs}
\usepackage{cite}
\usepackage{comment}
\usepackage{enumitem}
\usepackage{graphicx}
\usepackage{listings}
\usepackage{multirow}
\usepackage{tabularx}
\usepackage{textcomp}
\usepackage[most]{tcolorbox}
\usepackage{ulem}
\usepackage{url}
\usepackage{xcolor}
\usepackage{lipsum} 

\def\BibTeX{{\rm B\kern-.05em{\sc i\kern-.025em b}\kern-.08em
    T\kern-.1667em\lower.7ex\hbox{E}\kern-.125emX}}

\definecolor{vgreen}{RGB}{104,180,104}
\definecolor{vblue}{RGB}{49,49,255}
\definecolor{vorange}{RGB}{255,143,102}
\definecolor{comment}{rgb}{0,.6,0}
\definecolor{keyword}{rgb}{.63,0,.42}
\definecolor{kw2}{rgb}{.50,.50,.15}
\definecolor{kw3}{rgb}{.42,.42,.63}
\definecolor{string}{rgb}{1,0,0}

\lstdefinestyle{vcode}{
    language=Verilog,
    commentstyle=\color{comment},
    stringstyle=\color{string},
    keywordstyle=\bfseries\color{keyword},
    basicstyle=\footnotesize\ttfamily,
    captionpos=b,
    frame=single,
    rulecolor=\color{black}
    numbers=left,
    showstringspaces=false,
    tabsize=2,
    xleftmargin=2em
}

\newif\ifrev
\revtrue
\ifrev
  \newcommand{\xiaolong}[1]{{\color{red} [Xiaolong: #1]}}
\else  
  \newcommand{\xiaolong}[1]{}
\fi

\newif\ifrev
\revtrue
\ifrev
  \newcommand{\kaichen}[1]{{\color{green} [Kaichen: #1]}}
\else  
  \newcommand{\kaichen}[1]{}
\fi

\newif\ifrev
\revtrue
\ifrev
  \newcommand{\raj}[1]{{\color{blue} [Raj: #1]}}
\else  
  \newcommand{\raj}[1]{}
\fi

\author{
    \IEEEauthorblockN{Weimin Fu\IEEEauthorrefmark{1},
	Shijie Li\IEEEauthorrefmark{2}, 
	Yifang Zhao\IEEEauthorrefmark{2},
    Haocheng Ma\IEEEauthorrefmark{6},
	Raj Dutta\IEEEauthorrefmark{5},	
	Xuan Zhang\IEEEauthorrefmark{4}, 
	Kaichen Yang\IEEEauthorrefmark{3},		
	Yier Jin\IEEEauthorrefmark{2},
    Xiaolong Guo\IEEEauthorrefmark{1}
	}
        \IEEEauthorblockA{\IEEEauthorrefmark{1}Kansas State University, \{weiminf, guoxiaolong\}@ksu.edu}	
    \IEEEauthorblockA{\IEEEauthorrefmark{2}University of Science and Technology of China, \{shijie\_li, zhaoyifang\}@mail.ustc.edu.cn, jinyier@ustc.edu.cn}
    \IEEEauthorblockA{\IEEEauthorrefmark{3}Michigan Technological University
     kaicheny@mtu.edu}
    \IEEEauthorblockA{\IEEEauthorrefmark{4}Washington University in St. Louis
     xuan.zhang@wustl.edu}
    \IEEEauthorblockA{\IEEEauthorrefmark{5}Silicon Assurance
     rajgautamdutta@siliconassurance.com}  
     \IEEEauthorblockA{\IEEEauthorrefmark{6}Tianjin University,
     hc\_ma@tju.edu.cn} 
}
\title{\Large\bf Hardware Phi-1.5B: A Large Language Model Encodes Hardware Domain Specific Knowledge}
\IEEEspecialpapernotice{Invited Paper}

\begin{document}
\maketitle

\begin{abstract}
In the rapidly evolving semiconductor industry, where research, design, verification, and manufacturing are intricately linked, the potential of Large Language Models to revolutionize hardware design and security verification is immense. The primary challenge, however, lies in the complexity of hardware-specific issues that are not adequately addressed by the natural language or software code knowledge typically acquired during the pretraining stage. Additionally, the scarcity of datasets specific to the hardware domain poses a significant hurdle in developing a foundational model. Addressing these challenges, this paper introduces \textit{Hardware Phi-1.5B}, an innovative large language model specifically tailored for the hardware domain of the semiconductor industry. We have developed a specialized, tiered dataset—comprising small, medium, and large subsets—and focused our efforts on pre-training using the medium dataset. This approach harnesses the compact yet efficient architecture of the Phi-1.5B model. The creation of this first pre-trained, hardware domain-specific large language model marks a significant advancement, offering improved performance in hardware design and verification tasks and illustrating a promising path forward for AI applications in the semiconductor sector.
\end{abstract}


\section{Introduction}

Knowledge and language processing are crucial in multiple critical aspects of the semiconductor industry, such as research, design, verification, and manufacturing. These aspects involve complex interactions among numerous entities, demanding high accuracy and efficiency in information exchange. Artificial intelligence (AI) has demonstrated significant potential in fields such as hardware design, verification, and routing in recent years~\cite{hains2018towards,lai2023chipformer,cheng2021joint,mirhoseini2021graph, CircuitTraining2021}. Nevertheless, it should be noted that these AI applications do not yet fully exploit the entirety of available knowledge or information, leading to errors in their judgments based on partial data, relegating AI methodologies to a supplementary role in the hardware domain. In contrast, Large Language Models (LLMs) can extensively leverage the knowledge conveyed and utilized through natural language and code during the hardware design process. Given these capacities, LLMs possess the potential to revolutionize the fields of hardware design and security verification.

Fig.~\ref{fig:llmstage} illustrates the progression of stages in training LLMs from raw datasets to assistants. In the hardware domain, research endeavors are presently focused on In-Context Learning (ICL) strategies, such as hardware bug fixing and verification assistant~\cite{ahmad2023fixing,zhang2023llm4dv,kande2023llmassisted}. ICL does not alter parameters in LLMs; instead, it serves more as a tactical approach than a fundamental solution for performance enhancement. Consequently, irrespective of how the initial prompts are optimized, the improvements in model performance cannot be directly attributed to ICL. Supervised Finetuning (SFT) represents another stage. Several groups attempted to employ SFT to address hardware debugging and design challenges~\cite{fu2023llm4sechw, thakur2023benchmarking,thakur2023verigen}, but outcomes often lack consistency. The primary challenge lies in the complexity inherent to hardware-specific issues not adequately addressed by the natural language or software knowledge acquired during the Pretrain stage. Hence, developing a base model tailored to enhance robustness in the hardware domain is in high demand and will significantly strengthen the open-source hardware community.

However, the scarcity of datasets in the hardware domain presents the initial challenge in Pretrain stage. This scarcity is not unique to the hardware domain; datasets have become the most limited yet critical resource in developing LLMs. Leading proprietary LLM GPT-4~\cite{openai2023gpt4}, as well as the most potent open-source LLM Llama2~\cite{touvron2023llama}, do not furnish datasets. However, they share methodologies or \textit{recipes} for Pretrain dataset construction. Informed by the approaches delineated in constructing open-source datasets RedPajama~\cite{together2023redpajama} and the Stack~\cite{Kocetkov2022TheStack}, based on these recipes, we have crafted a Hardware Domain-Specific Dataset. This dataset has been segmented into three distinct tiers based on content volume: small, medium, and large.

In this study, we have adopted the Phi-1.5B model~\cite{textbooks2} architecture and conducted pretraining of a hardware domain-specific LLM on the medium dataset\footnote{Our available computational resources informed this decision.}. The Phi-1.5B model boasts performance that rivals that of Llama2 7B despite being only a fifth of its size, an attribute that underscores the model’s efficiency and effectiveness in training. Moreover, the reduced scale of the model also implies that the training costs for subsequent task-specific fine-tuning will be significantly lower. We anticipate that this fully open-source pretrained model will be a robust foundational support for a wide array of tasks within the hardware domain, especially playing a pivotal role in addressing hardware security challenges.
Through meticulously constructed hardware-specific datasets and customized pre-trained LLMs, this paper aims to precisely address the unique challenges in hardware domain tasks, achieving a deep understanding and response to the needs of this field. Our contributions are mainly reflected in the following aspects:

\begin{figure}[!t]
    \centering
    \includegraphics[width=\linewidth]{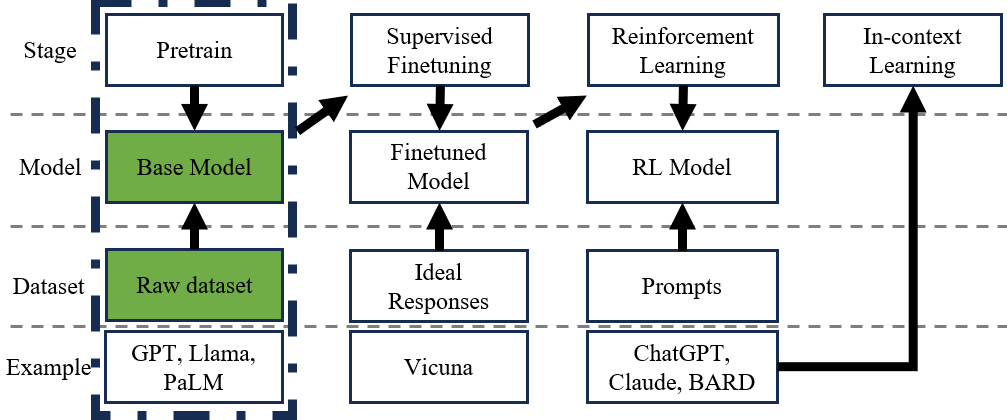}
    \caption{Four-Stage LLM-Based Assistant Development: Pretraining with raw data for a base model; ideal response-driven supervised fine-tuning; instruction-based reinforcement learning with few-shot examples for a deployable model; culminating in user engagement via in-context learning. The green cells highlight the contributions made in this paper.}
    \label{fig:llmstage}
\end{figure}

\begin{enumerate}[left=0pt]
    \item This paper conducted pretraining based on the Phi-1.5B model structure, making it more closely aligned with the needs of the hardware domain, enhancing the model's performance and stability in hardware design and verification tasks. To our knowledge, it is the first pretrained hardware domain-specific LLM. 
    \item  We created three differently sized datasets rigorously screened and optimized them to guarantee content relevance and quality, thus laying a strong foundation for model training.
    \item The pre-trained model is offered openly to the community, thus supporting ongoing research, development, and innovation in both academic and industrial spheres.

\end{enumerate}

\section{Background: Foundational Concepts}\label{sec:background}
\begin{figure}[!t]
    \centering
    \includegraphics[width = \columnwidth]{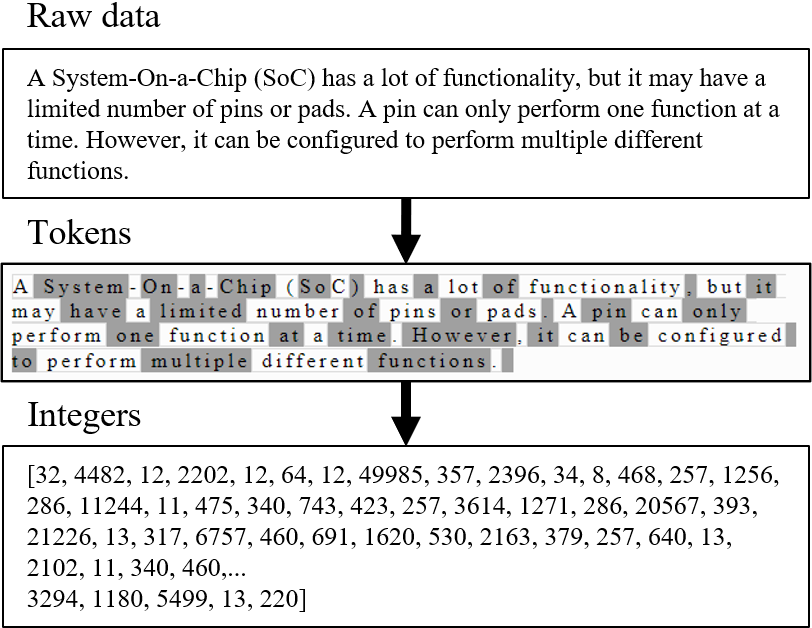}
    \caption{Tokenization example: transform all text to a list of integers.}
    \label{fig: tokenization}
\end{figure}
LLMs are not inherently capable of directly processing raw knowledge and information. To make the information intelligible to these models, we must first transform it into a sequence of integers that the model can directly interpret. This transformation process is known as \textbf{Tokenization}. As illustrated in Fig.~\ref{fig: tokenization}, we employ the CodeGen-mono~\cite{nijkamp2022codegen} tokenizer to segment the raw text into discrete units, referred to as \textit{tokens}. Note that although a tokenizer's vocabulary may contain a vast array of words, a token does not always correspond directly to a single word. For instance, the proprietary term \textit{SoC} depicted in the figure is split into two separate tokens: \textit{So} and \textit{C}. Each token is subsequently assigned a unique numerical identifier correlating to an index in the vocabulary list, generating an extensive sequence of integers. Upon analyzing our dataset, we observed that, on average, each token is equivalent to $0.74$ words.

\begin{figure}[!ht]
    \centering
    \includegraphics[width = \columnwidth]{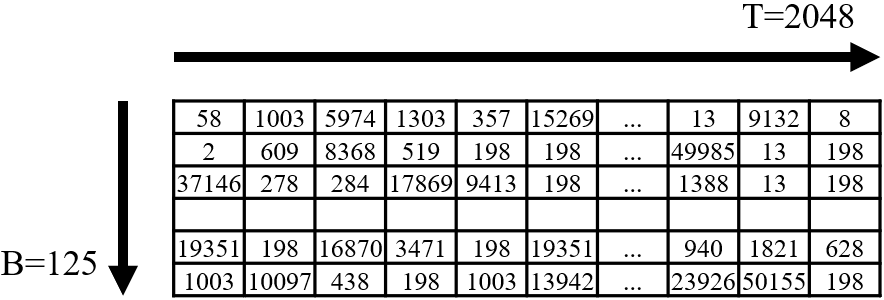}
    \caption{Matrix representation of the batch structure in Hardware Phi-1.5B.}
    \label{fig: batch}
\end{figure}

In the initial data handling phase, we encounter data characterized by elements of disparate lengths and a discrete nature. To methodologically address this heterogeneity, we used the batch structure as depicted in Fig.~\ref{fig: batch}, which is organized into a matrix with dimensions $(B, T)$. Here, $B$ denotes the batch size—fixed at $125$ for the scope of our experimental analysis—and $T$ encapsulates the maximal context length that our model, designated as Phi-1.5B, is capable of processing, the value of which is set at $2048$ tokens.

\textit{Batch Size} is the number of training examples utilized in one iteration. A vast batch size may exceed the memory constraints, precipitating out-of-memory errors. Conversely, an unduly small batch size could result in excessively noisy gradient updates, thereby detrimentally affecting model performance. For comparative context, the Llama2 employs a batch size $64$, correlating to a $64 \times 4096$ configuration. In contrast, the original MicroSoft Phi-1.5B utilizes a batch set at $2048 \times 2048$.

We employ a concatenation strategy to account for the inherent variability in the lengths of data items within a dataset—which is unlikely to match the specified value of $T$ uniformly. This involves unifying disparate data lengths and affixing an end-of-sequence (EOS) token to the culmination of each data item. Subsequently, the data undergoes a normalization process, uniformly resized to conform to the dimensions $(B, T)$. In our research, the EOS token is denoted by the symbol \texttt{<|endoftext|>}.

\begin{figure}[!ht]
    \centering
    \includegraphics[width=\linewidth]{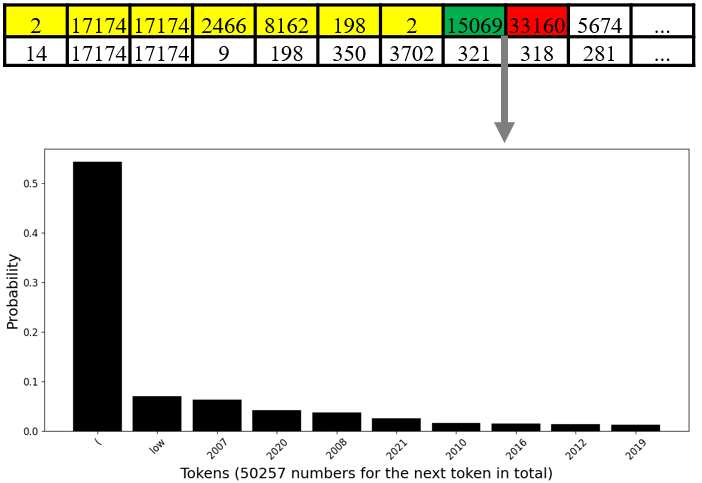}
    \caption{Token prediction probability distribution with preferences after pretrain.}
    \label{fig: predict}
\end{figure}

Fig.~\ref{fig: predict} illustrates the pretraining process of the LLMs, explicitly highlighting their capability to predict the next token. Within each cell of the training batch, the model is only privy to the data in the cells to the left within the same row. During the pretraining, the model learns to predict the content of the right-side cell by utilizing the context provided by the cells on the left, a paradigm known as a causal language model (CLM). The green cell represents a randomly selected token awaiting processing, the yellow area denotes the preceding text utilized for prediction, and the red cell signifies the prediction target. \textit{Hardware Phi-1.5B} would select from the entire vocabulary list comprising $50,257$ options. Through training, the model transitions from uniform probability distribution across all possible outcomes to allocating probabilities with a clear preference toward a handful of potential choices.

\section{Pretrain Dataset}\label{sec:method}

\subsection{Dataset Overview}

In this study, we have developed a hardware domain-specific dataset derived from various publicly accessible sources and bifurcated it into two main categories: code and natural language. Table~\ref{tab: Dataset_Construction} delineates the primary constituents of the dataset.

\begin{table}[!t]
\centering
\caption{Overview of Dataset Composition: Categories, Languages, Sources, and Selection Methods}
\label{tab: Dataset_Construction}
\resizebox{\columnwidth}{!}{%
\begin{tabular}{|c|c|c|c|}
\hline
\textbf{Data Category} & \textbf{Languages} & \textbf{Sources} & \textbf{Methodology or Selection Criteria} \\ \hline
\begin{tabular}[c]{@{}c@{}}Hardware Design\\ Code\end{tabular} & \begin{tabular}[c]{@{}c@{}}SystemVerilog, \\ Verilog, VHDL\end{tabular} & \begin{tabular}[c]{@{}c@{}}Google BigQuery Github,\\ the Stack\end{tabular} & \begin{tabular}[c]{@{}c@{}}Filtering GitHub repositories \\ for hardware design code\end{tabular} \\ \hline
\begin{tabular}[c]{@{}c@{}}Hardware Nature \\ Language Knowledge\end{tabular} & English & \begin{tabular}[c]{@{}c@{}}RedPajama CommmolCrawl,\\ ArXiv, StackExange, Books,\\ C4, Wikipedia), TrustHub, \\ Cad4Assurances, CWE\end{tabular} & \begin{tabular}[c]{@{}c@{}}Aggregating content from the \\ hardware domain\end{tabular} \\ \hline
\end{tabular}%
}
\end{table}

For the code segment, we leveraged Google BigQuery GitHub Public Datasets~\cite{hoffa2016github}
, selecting projects that encompass hardware design source code. Our selection was concentrated on three pivotal hardware programming languages: SystemVerilog, Verilog, and VHDL. This concentration facilitated the incorporation of entire repositories containing pertinent code into our dataset. Additionally, to ensure both legal compliance and open accessibility, we meticulously selected projects under specific open-source licenses.

In parallel, we curated a comprehensive hardware security dataset, amalgamating both code and natural language content from eminent hardware security sources, including TrustHub~\cite{TrustHub}, CAD for Assurance of Electronic Systems~\cite{cadforassurance}
, and Common Weakness Enumeration (CWE)~\cite{cwe2022}
. This compilation enriches our dataset with current and vital insights into hardware security, encompassing codes, their descriptions, best practices, and security recommendations for hardware design.

For practical applications in Code Language Models, integrating natural language data is commonly recognized as beneficial for enhancing model performance~\cite{roziere2023code}. Hence, we adopted and modified the Redpajama dataset construction methodology with a focused segmentation due to the relative rarity of hardware-specific content. The Redpajama dataset, an expansive corpus exceeding 1.2 trillion tokens, was segmented to enhance hardware-related discourse. The first segment includes data from specialized platforms such as Arxiv 
, Books
, Wikipedia
, and StackExchange
; the second segment is derived from broader internet content via CommonCrawl 
 and C4
.

Table~\ref{tab: DataFiltering} provides a comprehensive overview of the rigorous processes applied to verify and cleanse our dataset, ensuring its quality and integrity. These steps are crucial for maintaining the dataset’s reliability and usability in research:
\begin{itemize}[left=0pt]
    \item \textit{Verification and Cleansing}: This step involves identifying syntactic errors and reviewing natural language descriptions to ensure data accuracy. It combines automated scripts for efficiency and manual reviews for precision, addressing the dual aspects of mechanical accuracy and contextual relevance.
    \item \textit{Redpajama Dataset Filtering:} Here, we filter hardware security content using targeted keywords. This step is vital for maintaining the dataset's focus and relevance to the field of hardware security, ensuring that the dataset remains aligned with its intended purpose.

    \item \textit{Text Processing:} This phase includes several processes: NFC normalization, short content filtering, and removing punctuation and unnecessary spaces. The use of datasketch~\cite{zhu2023datasketch}, a tool known for its efficiency in handling large datasets, helps streamline this process, improving data quality and consistency.

    \item \textit{De-duplication:} To enhance the dataset's utility, this step employs indexing and the MinHash technique for de-duplication, again utilizing datasketch. De-duplication is critical for eliminating redundant data, enhancing overall quality, and making the dataset more manageable and effective for users.

\end{itemize}
Each of these steps plays a vital role in refining the dataset. This meticulous process ensures that the dataset is extensive but also precise and reliable, catering to the nuanced needs of hardware security research.

\begin{table}[!t]
\centering
\caption{Steps in Data Verification, Cleansing, and Processing with Applied Tools and Methods}
\label{tab: DataFiltering}
\resizebox{\columnwidth}{!}{%
\begin{tabular}{|c|c|c|}
\hline
\textbf{Step} & \textbf{Description} & \textbf{Tools/Methods} \\ \hline
Verification \& Cleansing & \begin{tabular}[c]{@{}c@{}}Identifying syntactic errors; \\ Reviewing natural language descriptions.\end{tabular} & \begin{tabular}[c]{@{}c@{}}Automated scripts,\\ Manual  review\end{tabular} \\ \hline
Redpajama Dataset Filtering & \begin{tabular}[c]{@{}c@{}}Filtering content related to \\ hardware using keywords\end{tabular} & Keyword filtering \\ \hline
Text Processing & \begin{tabular}[c]{@{}c@{}}NFC normalization; Filtering short content; \\ Removing punctuation, spaces, etc.\end{tabular} & datasketch \\ \hline
De-duplication & \begin{tabular}[c]{@{}c@{}}Indexing and de-duplication \\ using MinHash\end{tabular} & datasketch \\ \hline
\end{tabular}%
}
\end{table}

\subsection{Visual and Quantitative Analysis}

\begin{figure}
    \centering
    \includegraphics[width = \columnwidth]{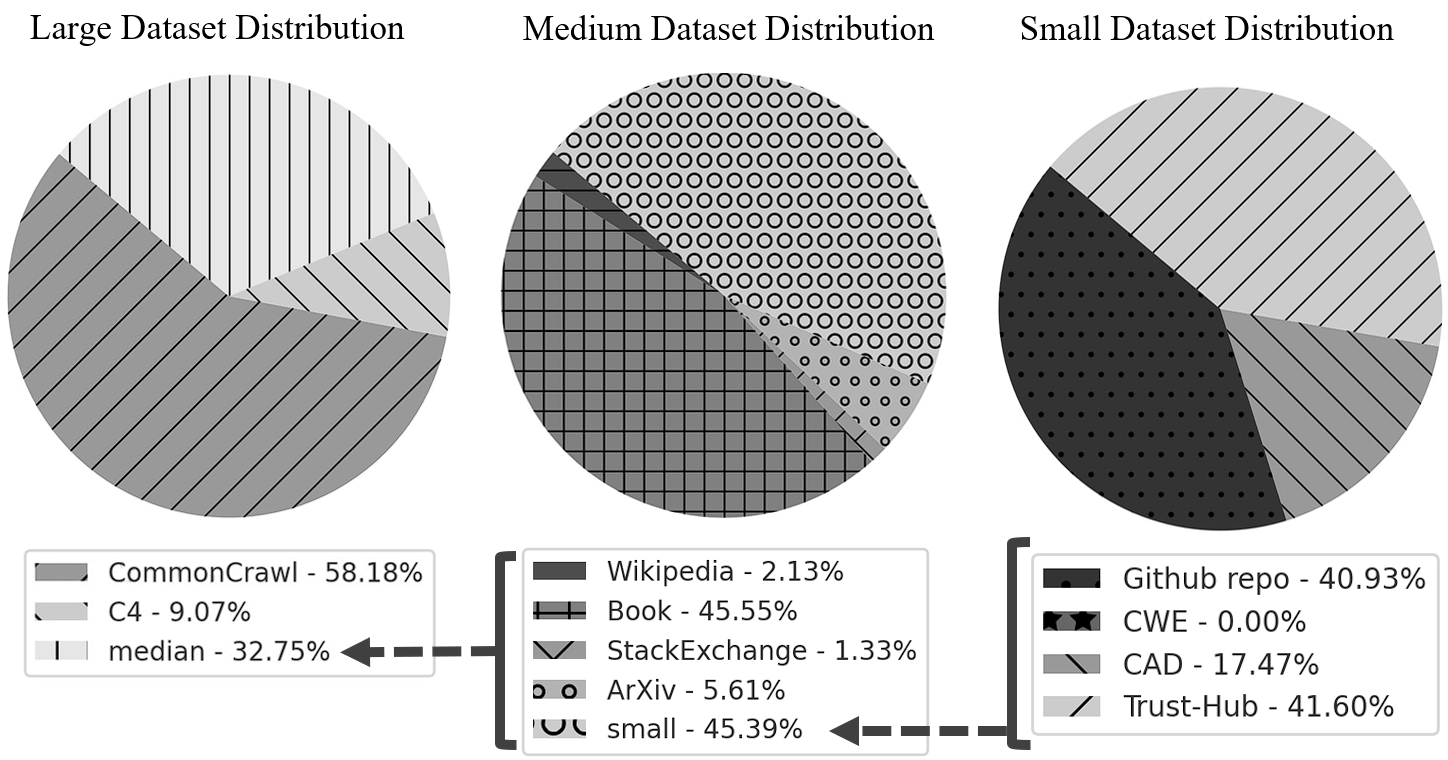}
    \caption{Visual Representation of Dataset Construction and Segmentation}
    \label{fig:datasetvisual}
\end{figure}

Fig.~\ref{fig:datasetvisual} visually represents our dataset's construction and segmentation. The smallest dataset comprises information related to hardware security and the source code of hardware design. In this dataset, the proportion of CWE is approximately $0.0014468\%$, yet its significance is paramount. This CWE section contains over $70,000$ tokens. Although it represents a minor proportion of the dataset's total volume, CWE information is crucial, offering critical insights into security vulnerabilities and weaknesses relevant to hardware security. While numerically limited, these data are of high quality and professionalism, reflecting common security weaknesses and potential risks in hardware design. Hence, despite their small percentage in the overall dataset, the CWE elements are indispensable and significantly contribute to in-depth hardware security research.

Expanding to the second dataset, the medium dataset evolves from the smaller dataset and incorporates information from ArXiv, StackExchange, Books, and Wikipedia sources. This dataset provides a broader perspective and information base, supporting a more comprehensive analysis.

Finally, the large dataset further extends the scope, encompassing all contents of the medium dataset and a vast array of data from C4 and CommonCrawl. The diversity and scale of this dataset offer abundant resources.

To further quantify, Table~\ref{tab:composition} presents a detailed breakdown of the three datasets in size, measured by token counts, facilitating a direct comparison between the small, medium, and large datasets.  This information is critical for users in selecting the most appropriate subset for their research needs. As an open-source dataset, this flexibility empowers users to select a subset that aligns with their computational resources and specific requirements.

\begin{table}[!t]
\centering
\caption{Breakdown of Datasets by Size in Tokens}
\label{tab:composition}
\begin{tabular}{|c|c|c|}
\hline
\textbf{Dataset Name} & \textbf{Dataset Size (Tokens)} \\ \hline
Smalt & $4,838,384,488$ \\ \hline
Medium  & $10,382,663,651$ \\ \hline
Large & $22,616,170,041$ \\ \hline
\end{tabular}%
\end{table}

\section{Hardware Phi-1.5B Model Pretrain}\label{sec:method_train}

\subsection{Model Architecture}
The architecture of Hardware Phi-1.5B strictly adheres to the design principles of the original Phi-1.5B and its variants. It comprises a Transformer structure~\cite{vaswani2017attention} with $24$ layers, $32$ heads, and a dimension of $64$ for each head, resulting in a context length of $2,048$. We incorporated the \textit{Flash Attention 2}~\cite{dao2022flashattention} during the training phase to expedite the training process. 
To ensure optimal compatibility with tools that utilize LLMs, we opted to follow the Phi-1.5B style and employ the codegen-moni tokenizer.

\subsection{Model Training Methodology}
The key setups in pretraining Hardware Phi-1.5B are listed below.

\subsubsection{Initialization Strategy}
The model's training is initiated with a state of random weights. We meticulously initialized the weights of the linear and embedding layers by employing a normal distribution with a mean of $0$ and a standard deviation of $0.02$. This approach was adopted to prevent the extremes of weight magnitude, thus averting the well-known issues of gradient disappearance or explosion. Additionally, we set the biases in linear layers to zero, fostering a neutral starting point that prevents any early bias toward specific outcomes.

\subsubsection{Training Configurations}
Our training configuration was standardized with a fixed learning rate of $2e-4$ and a weight decay factor of $0.1$, mirroring the training regimen of the Phi-1.5B model. 

\subsubsection{Optimizer Settings}
The Adam optimizer~\cite{kingma2014adam}, equipped with beta momentum values of $0.9$ and $0.98$ and an epsilon value of $1e-7$, was the chosen algorithm for its reliable performance in similar tasks. 

\subsubsection{Efficiency Strategies}
To optimize memory usage and training efficiency, we adopted fp16 mixed precision training. Additionally, we utilized the Fully Sharded Data Parallel~\cite{xu2020automatic}, enabling the distribution of the model's parameters across all available GPU resources. This was complemented by an effective communication strategy aimed at reducing training overheads. 
For further memory optimization, the transformer blocks were integrated with the \texttt{auto\_wrap\_policy}, and to strike a balance between memory use and computation speed, we enabled activation checkpointing.

\subsubsection{Evaluation Framework}
Given the substantial size of our dataset, we set a training termination criterion at $750,000$ iterations and $30,000$ steps. To monitor the model's performance progression and mitigate the potential of overfitting, we instituted a checkpoint mechanism that allowed the model's state to be saved and evaluated at every $1,000$ step, ensuring we could capture performance metrics systematically throughout the training process.

\section{Experiment and Result}\label{sec:expr}
Our training platform is constructed on a server operating with Ubuntu 20.04.6 LTS, equipped with an Intel(R) Xeon(R) Silver 4314 CPU ($2.40$ GHz, $64$ cores), $251$ GB of memory, and dual NVIDIA A100 $80$ GB graphics processing cards. This configuration not only provides substantial computational power but also ensures ample memory when dealing with large models. The CUDA 12.2 and PyTorch 2.1.0 versions selected for our work fully exploit the hardware potential, optimizing the model training process.
Supported by this high-performance hardware, our platform can achieve a throughput of $1.07$ batches per second while maintaining approximately $100T$ floating-point operations per second (flops/sec) and a token processing speed of around $11k$ per second.

In this experiment, we have designated $30k$ training steps, culminating in a total training duration of $8$ days, $2$ hours, $43$ minutes, and $22$ seconds. This training cycle was determined post-evaluation of the anticipated model complexity and the requisite time for convergence. Additionally, considering the power consumption of GPU devices and based on previous studies~\cite{bender2021dangers}
, we have computed the energy efficiency during the training process. We estimate that this training has produced approximately $90$kg of carbon dioxide equivalent greenhouse gas emissions.

\begin{figure}[!ht]
\centering
\includegraphics[width=\columnwidth]{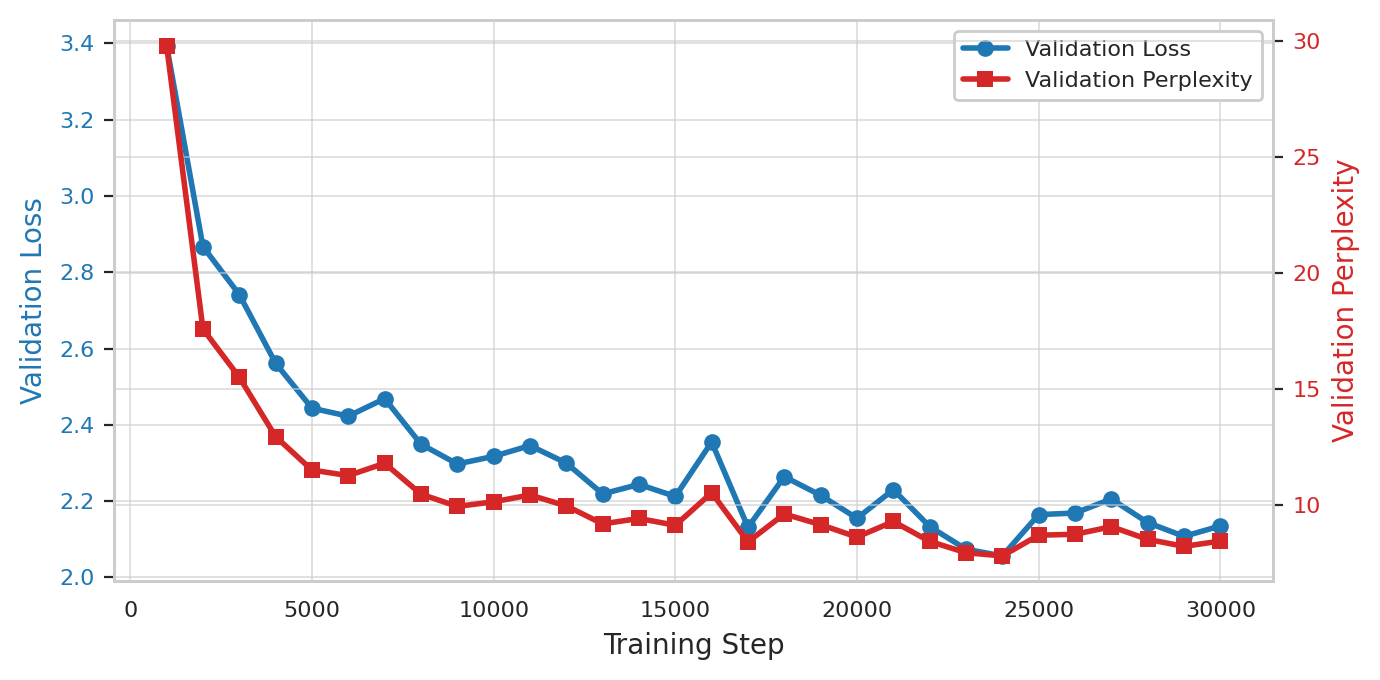}
\caption{Validation Loss and Perplexity During Training}
\label{fig:val_loss_ppl}
\end{figure}

Fig.~\ref{fig:val_loss_ppl} presents the variation in loss and perplexity on the validation dataset. Loss is an indicator that measures the discrepancy between model predictions and actual values, with Mean Squared Error (MSE) and Cross-Entropy Loss being commonly used metrics~\cite{goodfellow2016deep}. A high loss indicates a greater disparity between the model predictions and actual values, hence a lower model performance. Perplexity~\cite{manning1999foundations}, particularly within natural language processing and language models, indicates model performance. It is the exponentiation of the cross-entropy loss and offers an average branching factor per word, that is, the model's uncertainty about the next word given the preceding ones. In our training context, a lower perplexity suggests a more precise understanding of the data by the model.

Initially, models exhibit high loss and perplexity, reflecting their preliminary adaptation to the data during the learning process. As training progresses, these metrics generally decrease, signifying the model's advancements in learning.
As in Fig.~\ref{fig:CWE}, we have conducted text continuation experiments on the CWE-1189~\cite{cwe1189} security vulnerability description.
In our experiments, we generated text using models at different training stages—the initial model, the model after $10k$ steps, the model after $20k$ steps, and the final model at $30k$ steps training.

\begin{figure}
    \centering
    \fbox{%
        \begin{minipage}{8cm}
        A System-On-a-Chip (SoC) has a lot of functionality, but it may have a limited number of pins or pads. A pin can only perform one function at a time. However, it can be configured to perform multiple different functions. This technique is called pin multiplexing.
        \end{minipage}
    }
    \caption{Training Data: CWE-1189: Improper Isolation of Shared Resources on System-on-a-Chip (SoC)}
    \label{fig:CWE}
\end{figure}

In Fig.~\ref{fig:init}, the initial model's generated text lacked coherence and logic, indicating that at the onset of training, the model's understanding of language structure had yet to be established. This stage's model output is typically a meaningless string of words, reflecting the model's randomness and immaturity.

\begin{figure}
    \centering
    \fbox{%
        \begin{minipage}{8cm}
        tributePAfort MoorFontSizeestone Karl Norris intens Norris informative refinement kg 278 impulses Yugoslavinnamon Buddha Performance Divine clinch interacts Performance Scientology alertinsulaes inex Harding Gene oral Norris heinous risk Kremlin Harding546 crossings 
        \end{minipage}
    }
    \caption{Generated sample at initialization.}
    \label{fig:init}
\end{figure}
After $10k$ training steps, the model demonstrated correct word spelling as in Fig.~\ref{fig:10k}, signifying that it had begun to capture the basic language rules, such as word boundaries and fundamental grammatical structures. However, the generation of longer sentences remained unnatural, suggesting an insufficient grasp of complex structures.
\begin{figure}
    \centering
    \fbox{
        \begin{minipage}{8cm}
        if 0 = Pin multiplexing else do you need to change those pin numbers to the least expensive chips? A: The pin number of the SoC has to be the same in both the input and output pins, with no extra pins on the chip connected directly. There are several sources of noise, either from the FPGA or a FPGA or other components like these may be bad quality. 
        \end{minipage}
    }
    \caption{Generated sample at $10k$ step, $250k$ iterations of training.}
    \label{fig:10k}
\end{figure}

Following $20k$ training steps, as shown in Fig.~\ref{fig:20k}, the model's generated text began to show readability, albeit with occurrences of looping and repetition. This may be attributed to the model finding a "comfort zone" in certain local minima and tending to repeat patterns discovered in the training data rather than generating novel content.
\begin{figure}
    \centering
    \fbox{%
        \begin{minipage}{8cm}
        On-board: I/O multiplexing
I/O multiplexing
For example:
I/O multiplexing
I/O multiplexing
I/O multiplexed in between
As mentioned in the link, a physical system-on-a-chip (SoC), also known as a Chip-on-a-Chip (SoC), offers pinout, multiplexer and multiplexer connected in a single interconnect. 
        \end{minipage}
    }
    \caption{Generated sample at $20k$ steps, $500k$ iterations of training.}
    \label{fig:20k}
\end{figure}

Finally, the model post $30k$ training steps exhibited a more mature understanding of language, integrating natural language capability and domain-specific knowledge in text generation in Fig.~\ref{fig:30k}. Nonetheless, due to the lack of task-specific fine-tuning, the model might not adapt well to specific contexts or answer complex queries. This highlights the model's limitations as a base model: it is adept at predicting the next token in text and, hence, tends to complete a document, but for more advanced tasks, further optimization and adaptation may be necessary.
\begin{figure}
    \centering
    \fbox{%
        \begin{minipage}{8cm}
        In the context of embedded-edge technologies, in the processor, an edge device is an integrated hardware device. An edge device is an embedded computation device that provides the capability to operate at all the levels of the chip. In the context of processors, it is more common that an embedded computation device is integrated and integrated into an embedded system. In the above example, for an embedded computation device, it is referred to the processor as computing core. It also refers to the SoC device when it is the peripheral, embedded system, or embedded system controller. On the other hand, an embedded system is an embedded systems management device that provides the capability to communicate with the
        \end{minipage}
    }
    \caption{Generated sample at $30k$ steps, $750k$ iterations of training.}
    \label{fig:30k}
\end{figure}

Through these phases, we can observe the model's progression in understanding and generating language. Nevertheless, to become a practical assistant, the model requires further training and customized fine-tuning. Future work will focus on enhancing the model's performance while reducing its training process's environmental impact.

\section{Related Work}\label{sec:rela}
Compared to the singular pursuit of developing Large Language Models for the goal of achieving General Artificial Intelligence, an increasing body of research is focusing on constructing specialized, domain-specific datasets and training LLMs to nurture systems that demonstrate expert-level proficiency within specific domains. 

In the medical field, due to the highly complex nature of the expertise required and the constraints imposed by privacy regulations, general LLM typically fails to provide sufficient comprehensive coverage. As a result, researchers are turning to strategies involving complete pretraining, supervised finetuning, and reinforcement learning approaches. K. Singhal has developed MultiMedQA, a composite benchmark integrating six existing medical question-answering datasets, and a new online search medical question dataset, HealthSearchQA. Utilizing this benchmark, Google has further trained PaLM and its variant, FLAN-PaLM~\cite{singhal2022large}. L. Y. Jiang has adopted a BERT-based pretraining and finetuning to develop NYUTron, designed to offer guidance at clinical care points~\cite{jiang2023health}. 

In the realm of hardware security, research approaches vary depending on the perspectives of researchers. Some contend that existing commercial general-purpose LLMs, such as ChatGPT, are sufficient to support formal verification tasks, relying on OpenAI's continuous enhancements of ChatGPT to improve the performance of their tools. For instance, M. Orenes-Vera has attempted to use ChatGPT for RTL formal verification~\cite{orenes2023rtl}. M. Chen, on the other hand, has employed codex, ChatGPT, and a Codegen variant fine-tuned for Verilog~\cite{thakur2023benchmarking} to generate assertions directly~\cite{kande2023llm}. However, other researchers argue that proprietary knowledge in the hardware domain necessitates custom dataset training for models. They believe that their research outcomes will be more pronounced as access to more domain-specific data and computational resources becomes abundant. For example, S. Thakur has finetuned for Verilog generation~\cite{thakur2023benchmarking}, and W. Fu has also finetuned an LLM for hardware debugging based on version control information~\cite{fu2023llm4sechw}.

On the other hand, there can be significant variances in the usage of specialized terminology between different domains, even evident in the software and hardware fields. For example, \textit{Port} typically refers to a communication interface in software engineering, whereas it denotes a physical interface on electronic devices in hardware design. \textit{Cache} in software signifies a temporary storage area to expedite data access. At the same time, in hardware, it might refer to a small-capacity, high-speed storage located between the CPU and main memory. Moreover, \textit{Pipeline} can represent a sequence of processing steps in software development, whereas, in hardware design, it indicates a specific technique for parallel data processing. Terms like \textit{Bus}, \textit{Driver}, \textit{Register}, and \textit{Core} also possess dual meanings; their conflation can lead to confusion and impair the model's understanding and predictions.

In light of this, our research endeavors are concentrated on the development and use of datasets tailored explicitly for the hardware domain for pretraining, and aiming to construct a base model that paves the way for breakthroughs in the meticulous finetuning of domain-specific models. We anticipate that this approach will significantly enhance the model's performance in comprehending and handling the aforementioned complex terminologies, surpassing the capabilities of existing general code language models (such as CodeLlama~\cite{roziere2023code}) and natural language models (such as BERT~\cite{devlin2018bert} and GPT-2~\cite{radford2019language}).

\section{Conclusion and Future Work}\label{sec:con}
In this study, we have explored the potential of Large Language Models advancing hardware design, Electronic Design Automation, and hardware security. Recognizing that hardware design shares certain formal similarities with natural language and software design yet diverges fundamentally in its core complexities, we have focused on developing a specialized LLM for the hardware domain, establishing a robust foundational dataset. This endeavor not only breaks new ground in conventional approaches to the hardware domain but also paves the way for novel perspectives in future research and applications.

Moving forward, we aim to continue advancing this project, focusing on pre-training the base model while maintaining and updating our dataset to ensure its relevance and contemporaneity. We eagerly anticipate the fine-tuning and application of this model in specific areas within the hardware design domain, particularly in addressing distinct design challenges and security issues. We believe our work will offer new insights and solutions for research and practice in hardware design, verification, and security, potentially catalyzing transformative changes in these fields.
\section*{\sc Acknowledgments}
Portions of this work were supported by the National Science Foundation (CCF-2019310, First Award Program of ARISE in EPSCoR 2148878).
\bibliographystyle{IEEEtran}
\bibliography{Xiaolong}
\end{document}